%% file: 0_main.tex
\documentclass{article}
\usepackage{style/unites}
\usepackage{XCharter}
\usepackage[scaled=1.1]{zlmtt} 
\usepackage{wrapfig}

\input{style/macros}

\begin{document}

\makeatletter
\def\blfootnote{\gdef\@thefnmark{}\@footnotetext}
\makeatother

\makeatletter
\pagestyle{fancy}
\fancyhf{}
\renewcommand{\headrulewidth}{1pt}
\chead{\small\bf \input{1_title}
}
\cfoot{\thepage}
\thispagestyle{fancy}
\makeatother

\makeatletter
\def\icmldate#1{\gdef\@icmldate{#1}}
\icmldate{\today}
\makeatother

\makeatletter
\fancypagestyle{fancytitlepage}{
  \fancyhead{}
  \lhead{\includegraphics[height=0.8cm]{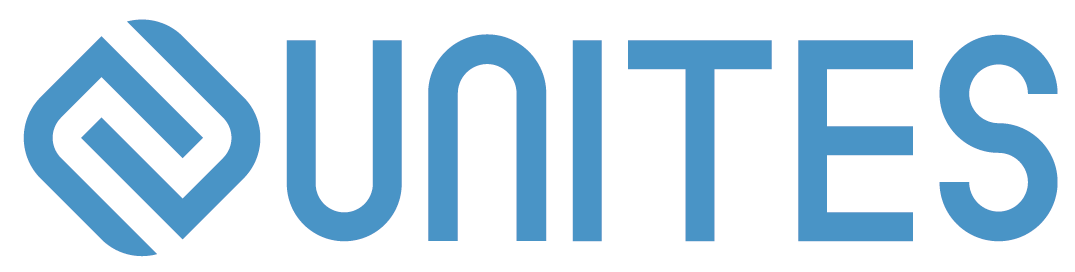}}
  \rhead{\it \@icmldate}
  \cfoot{}
}
\makeatother

\thispagestyle{fancytitlepage}

\vspace*{0.5em}

\noindent
\begin{titleblock}
    {\setlength{\parskip}{0cm}
     \raggedright
     {\setstretch{1.2}
      \fontsize{15}{18}\selectfont\bfseries
      \input{1_title}
      \par}
    }
    \vskip 0.2cm
    
    \input{2_authors}
    \vskip 0.2cm
    
    \input{tex/0_abs}
    
\end{titleblock}

\blfootnote{%
\ifcsname @icmlpreprint\endcsname
  \textit{\csname @icmlpreprint\endcsname}%
\fi
}

\input{tex/1_intro}

\input{tex/2_related.tex}
\input{tex/3_method}
\input{tex/4_experiments}

\input{tex/7_conclusion}
\input{99_acknowledgement}

\onecolumn
\bibliography{999_reference}
\bibliographystyle{plainnat}

\titlespacing*{\section}{0pt}{*1}{*1}
\titlespacing*{\subsection}{0pt}{*1.25}{*1.25}
\titlespacing*{\subsubsection}{0pt}{*1.5}{*1.5}

\setlength{\abovedisplayskip}{\baselineskip} 
\setlength{\abovedisplayshortskip}{0.5\baselineskip} 
\setlength{\belowdisplayskip}{\baselineskip}
\setlength{\belowdisplayshortskip}{0.5\baselineskip}

\clearpage
\appendix
\label{sec:append}
\part*{Appendix}
{
\setlength{\parskip}{-0em}
\startcontents[sections]
\printcontents[sections]{ }{1}{}
}

\setlength{\parskip}{.5em}
\input{tex/5_appendix}

\end{document}

%% file: style/macros.tex
\usepackage[utf8]{inputenc}
\usepackage[T1]{fontenc}
\usepackage{microtype}

\usepackage{amsmath}
\usepackage{amssymb}
\usepackage{amsfonts}
\usepackage{amsthm}
\usepackage{mathtools}
\usepackage{mathrsfs}
\usepackage{physics}
\usepackage{braket}
\usepackage{slashed}
\usepackage{nicefrac}
\usepackage{textcomp}
\usepackage{dsfont}
\usepackage{bbm}
\usepackage{bm}

\usepackage{graphicx}
\usepackage{subcaption}
\usepackage[export]{adjustbox}
\usepackage{float}
\usepackage{booktabs}
\usepackage{dcolumn}
\newcolumntype{d}[1]{D{.}{.}{#1}}
\usepackage{bigstrut, tabularx, multirow, makecell, diagbox}
\usepackage{colortbl}
\usepackage{tabularray}
\UseTblrLibrary{booktabs}
\usepackage{threeparttable}
\usepackage{tablefootnote}
\usepackage{fontawesome5}

\usepackage{placeins}
\usepackage{caption}
\usepackage{footnote}
\usepackage{enumitem}
\usepackage{multicol}
\usepackage{xspace}
\usepackage{titletoc}
\usepackage{titlesec}
\usepackage[bottom]{footmisc}
\usepackage{setspace}

\usepackage{wrapfig}
\usepackage{tikz}
\usepackage{quantikz}
\usepackage{dashbox}
\usepackage{mdframed}
\usepackage{marvosym}
\usepackage{pifont}
\usepackage{CJK}
\usepackage{url}

\usepackage[table,x11names]{xcolor}
\usepackage[most]{tcolorbox}
\tcbuselibrary{breakable}
\usetikzlibrary{decorations.pathreplacing, fit}

\definecolor{primaryblue}{HTML}{0066CC}
\definecolor{accentcyan}{HTML}{00D4AA}
\definecolor{warmorange}{HTML}{FF6B35}
\definecolor{deepgray}{HTML}{2C3E50}
\definecolor{lightgray}{HTML}{F8F9FA}
\definecolor{gradientstart}{HTML}{667eea}
\definecolor{gradientend}{HTML}{764ba2}

\definecolor{citecolor}{HTML}{0071bc}
\definecolor{citeblue}{RGB}{0, 113, 188}
\definecolor{linkcolor}{HTML}{9A4D92}
\definecolor{firebrick}{rgb}{0.698,0.133,0.133}

\definecolor{paleviolet}{HTML}{E1EEFC}
\definecolor{CarolinaUltraLight}{HTML}{E7F4FC}
\definecolor{lightgrey}{RGB}{247, 247, 247}
\definecolor{shadecolor}{HTML}{EFEFEF}
\definecolor{lightyellow}{rgb}{1.0, 0.95, 0.7}
\definecolor{lightblue}{rgb}{0.90, 0.95, 1.0}
\definecolor{light-gray}{gray}{0.95}

\definecolor{darkgrey}{rgb}{0.5, 0.5, 0.5}
\definecolor{darkgreen}{rgb}{0, 0.5, 0}
\definecolor{mydarkblue}{rgb}{0,0.08,0.45}
\definecolor{mydarkblue2}{rgb}{0.133, 0.133, 0.698}
\definecolor{echodrk}{HTML}{0099cc}
\definecolor{mymauve}{rgb}{0.58,0,0.82}
\definecolor{midnightblue}{rgb}{0.1,0.1,0.44}
\definecolor{oxfordblue}{rgb}{0.0,0.13,0.28}
\definecolor{prussianblue}{rgb}{0.0,0.19,0.33}
\definecolor{coolteal}{rgb}{0, 0.45, 0.45}
\definecolor{olive}{rgb}{0.1, 0.3, 0}
\definecolor{mypurple}{rgb}{0.5,0,0.5}
\definecolor{almond}{rgb}{0.94, 0.87, 0.8}

\definecolor{blue_ampEncoding}{HTML}{DAE8FC}
\definecolor{green_encoder}{HTML}{D5E8D4}
\definecolor{purple_decoder}{HTML}{E1D5E7}
\definecolor{yellow_measure}{HTML}{FFF2CC}
\definecolor{gray_block}{HTML}{F5F5F5}
\definecolor{pink_dru}{HTML}{FAD9D5}
\definecolor{orange_v}{HTML}{FAD7AC}

\definecolor{colorA}{rgb}{1,0,0}
\definecolor{colorB}{rgb}{0,0.3,1}
\definecolor{colorC}{rgb}{0.9,0.8,0.2}
\definecolor{colorD}{rgb}{0,0.65,0}
\definecolor{lesslightgray}{rgb}{0.5,0.5,0.5}
\definecolor{fundamental}{RGB}{55, 110, 111}
\definecolor{Gred}{RGB}{219, 50, 54}
\definecolor{ToCgreen}{RGB}{0, 128, 0}
\definecolor{Sepia}{RGB}{112, 66, 20}
\definecolor{Dblue}{rgb}{0,0.08,0.45}
\definecolor{Blue}{rgb}{0, 0, 0.8}
\definecolor{blue}{rgb}{0,0,1}
\definecolor{UNCblue!10}{rgb}{0.84,0.91,0.98}
\definecolor{RowAlt}{rgb}{0.98,0.98,0.99}

\definecolor{CarolinaBlue}{HTML}{7BAFD4}        
\definecolor{CarolinaLightBlue}{HTML}{B3D4E5}   
\definecolor{CarolinaUltraLight}{HTML}{E8F4F8}  
\definecolor{CarolinaText}{HTML}{1C2B33}        

\usepackage[pagebackref=true,breaklinks=true,colorlinks,hyperfootnotes=false]{hyperref}
\hypersetup{
  colorlinks,
  citecolor=citeblue,
  linkcolor=firebrick,
  urlcolor=firebrick
}
\usepackage[nameinlink,capitalize,noabbrev]{cleveref}

\titlespacing\section{0pt}{4pt plus 4pt minus 2pt}{-2pt plus 2pt minus 2pt}
\titlespacing\subsection{0pt}{2pt plus 4pt minus 2pt}{-2pt plus 2pt minus 2pt}
\titlespacing\subsubsection{0pt}{2pt plus 4pt minus 2pt}{-2pt plus 2pt minus 2pt}

\makeatletter
\def\th@remark{%
  \thm@headfont{\bfseries}%
  \normalfont 
  \thm@preskip\topsep \divide\thm@preskip\tw@
  \thm@postskip\thm@preskip
}
\makeatother

\theoremstyle{definition}

\tcolorboxenvironment{theorem}{
  breakable,
  colback=black!10,
  colframe=white,
  width=\linewidth, 
  enlarge left by=0pt,
  enlarge right by=0pt,
  boxsep=5pt,
  boxrule=0pt,
  left=0pt,right=0pt,top=0pt,bottom=0pt,
  arc=8pt,
  before skip=\topsep,
  after skip=\topsep
}

\tcolorboxenvironment{lemma}{
  breakable,
  colback=black!10,
  colframe=white,
  width=\linewidth,
  enlarge left by=0pt,
  enlarge right by=0pt,
  boxsep=5pt,
  boxrule=0pt,
  left=0pt,right=0pt,top=0pt,bottom=0pt,
  arc=8pt,
  before skip=\topsep,
  after skip=\topsep
}

\tcolorboxenvironment{corollary}{
  breakable,
  colback=black!10,
  colframe=white,
  width=\linewidth,
  enlarge left by=0pt,
  enlarge right by=0pt,
  boxsep=5pt,
  boxrule=0pt,
  left=0pt,right=0pt,top=0pt,bottom=0pt,
  arc=8pt,
  before skip=\topsep,
  after skip=\topsep
}

\tcolorboxenvironment{proposition}{
  breakable,
  colback=black!10,
  colframe=white,
  width=\linewidth,
  enlarge left by=0pt,
  enlarge right by=0pt,
  boxsep=5pt,
  boxrule=0pt,
  left=0pt,right=0pt,top=0pt,bottom=0pt,
  arc=8pt,
  before skip=\topsep,
  after skip=\topsep
}

\tcolorboxenvironment{definition}{
  breakable,
  colback=black!10,
  colframe=white,
  width=\linewidth,
  enlarge left by=0pt,
  enlarge right by=0pt,
  boxsep=5pt,
  boxrule=0pt,
  left=0pt,right=0pt,top=0pt,bottom=0pt,
  arc=8pt,
  before skip=\topsep,
  after skip=\topsep
}

\tcolorboxenvironment{assumption}{
  breakable,
  colback=black!10,
  colframe=white,
  width=\linewidth,
  enlarge left by=0pt,
  enlarge right by=0pt,
  boxsep=5pt,
  boxrule=0pt,
  left=0pt,right=0pt,top=0pt,bottom=0pt,
  arc=8pt,
  before skip=\topsep,
  after skip=\topsep
}

\tcolorboxenvironment{claim}{
  breakable,
  colback=black!10,
  colframe=white,
  width=\linewidth,
  enlarge left by=0pt,
  enlarge right by=0pt,
  boxsep=5pt,
  boxrule=0pt,
  left=0pt,right=0pt,top=0pt,bottom=0pt,
  arc=8pt,
  before skip=\topsep,
  after skip=\topsep
}

\tcolorboxenvironment{problem}{
  breakable,
  colback=black!10,
  colframe=white,
  width=\linewidth,
  enlarge left by=0pt,
  enlarge right by=0pt,
  boxsep=5pt,
  boxrule=0pt,
  left=0pt,right=0pt,top=0pt,bottom=0pt,
  arc=8pt,
  before skip=\topsep,
  after skip=\topsep
}

\tcolorboxenvironment{question}{
  breakable,
  colback=black!10,
  colframe=white,
  width=\linewidth,
  enlarge left by=0pt,
  enlarge right by=0pt,
  boxsep=5pt,
  boxrule=0pt,
  left=0pt,right=0pt,top=0pt,bottom=0pt,
  arc=8pt,
  before skip=\topsep,
  after skip=\topsep
}

\newtcolorbox{titleblock}{
  enhanced,
  frame hidden,
  colback=CarolinaUltraLight,
  colframe=CarolinaUltraLight,
  boxrule=0pt,
  arc=10pt,
  left=14pt,
  right=14pt,
  top=14pt,
  bottom=14pt,
  width=\linewidth,
  before skip=12pt plus 4pt,
  after skip=12pt plus 4pt,
  grow to left by=1.5pt,
  grow to right by=1.5pt,
  before upper={
    \setlength{\parindent}{0cm}
    \setlength{\parskip}{0.5cm}
  }
}

\crefname{theorem}{Theorem}{Theorems}
\crefname{proposition}{Proposition}{Propositions}
\crefname{lemma}{Lemma}{Lemmas}
\crefname{corollary}{Corollary}{Corollaries}
\crefname{definition}{Definition}{Definitions}
\crefname{assumption}{Assumption}{Assumptions}
\crefname{remark}{Remark}{Remarks}
\crefname{problem}{Problem}{Problems}
\crefname{property}{Property}{property}
\crefname{question}{Question}{Questions}

\numberwithin{equation}{section}
\numberwithin{theorem}{section}
\numberwithin{proposition}{section}
\numberwithin{definition}{section}
\numberwithin{lemma}{section}
\numberwithin{assumption}{section}
\numberwithin{remark}{section}

\def\1{\bm{1}}

\makeatletter
\let\save@mathaccent\mathaccent
\newcommand*\if@single[3]{%
    \setbox0\hbox{${\mathaccent"0362{#1}}^H$}%
    \setbox2\hbox{${\mathaccent"0362{\kern0pt#1}}^H$}%
    \ifdim\ht0=\ht2 #3\else #2\fi
}
\newcommand*\rel@kern[1]{\kern#1\dimexpr\macc@kerna}
\newcommand*\widebar[1]{\@ifnextchar^{{\wide@bar{#1}{0}}}{\wide@bar{#1}{1}}}
\newcommand*\wide@bar[2]{\if@single{#1}{\wide@bar@{#1}{#2}{1}}{\wide@bar@{#1}{#2}{2}}}
\newcommand*\wide@bar@[3]{%
    \begingroup
    \def\mathaccent##1##2{%
        \let\mathaccent\save@mathaccent
        \if#32 \let\macc@nucleus\first@char \fi
        \setbox\z@\hbox{$\macc@style{\macc@nucleus}_{}$}%
        \setbox\tw@\hbox{$\macc@style{\macc@nucleus}{}_{}$}%
        \dimen@\wd\tw@
        \advance\dimen@-\wd\z@
        \divide\dimen@ 3
        \@tempdima\wd\tw@
        \advance\@tempdima-\scriptspace
        \divide\@tempdima 10
        \advance\dimen@-\@tempdima
        \ifdim\dimen@>\z@ \dimen@0pt\fi
        \rel@kern{0.6}\kern-\dimen@
        \if#31
        \overline{\rel@kern{-0.6}\kern\dimen@\macc@nucleus\rel@kern{0.4}\kern\dimen@}%
        \advance\dimen@0.4\dimexpr\macc@kerna
        \let\final@kern#2%
        \ifdim\dimen@<\z@ \let\final@kern1\fi
        \if\final@kern1 \kern-\dimen@\fi
        \else
        \overline{\rel@kern{-0.6}\kern\dimen@#1}%
        \fi
    }%
    \macc@depth\@ne
    \let\math@bgroup\@empty \let\math@egroup\macc@set@skewchar
    \mathsurround\z@ \frozen@everymath{\mathgroup\macc@group\relax}%
    \macc@set@skewchar\relax
    \let\mathaccentV\macc@nested@a
    \if#31
    \macc@nested@a\relax111{#1}%
    \else
    \def\gobble@till@marker##1\endmarker{}%
    \futurelet\first@char\gobble@till@marker#1\endmarker
    \ifcat\noexpand\first@char A\else
    \def\first@char{}%
    \fi
    \macc@nested@a\relax111{\first@char}%
    \fi
    \endgroup
    }
\makeatother
\let\bar\widebar

\DeclareMathAlphabet{\mathsfit}{\encodingdefault}{\sfdefault}{m}{sl}
\SetMathAlphabet{\mathsfit}{bold}{\encodingdefault}{\sfdefault}{bx}{n}

\let\tilde\widetilde
\let\hat\widehat

\renewcommand{\arraystretch}{1.15}


%% file: 1_title.tex
Geometry- and Relation-Aware Diffusion for EEG Super-Resolution

%% file: 2_authors.tex
\begin{icmlauthorlist}
\mbox{Laura Yao$^{\,1\,*}$},
\mbox{Gengwei Zhang$^{\,1\,*}$}, 
\mbox{Moajjem Chowdhury$^{\,2\,}$}, 
\mbox{Yunmei Liu$^{\,2\,}$}, 
and \mbox{Tianlong Chen$^{\,1\,}$}
\end{icmlauthorlist}

$^{1\,}$UNITES Lab, University of North Carolina at Chapel Hill  \quad $^{2\,}$MINDxAI Lab, Dept. of Industrial \& Systems Engineering, University of Louisville, Louisville, KY, USA.

$^{*}$ Equal Contribution

%% file: tex/0_abs.tex
Recent electroencephalography (EEG) spatial super-resolution (SR) methods, while showing improved quality by either directly predicting missing signals from visible channels or adapting latent diffusion-based generative modeling to temporal data, often lack awareness of physiological spatial structure, thereby constraining spatial generation performance. To address this issue, we introduce TopoDiff, a geometry- and relation-aware diffusion model for EEG spatial super-resolution. Inspired by how human experts interpret spatial EEG patterns, TopoDiff incorporates topology-aware image embeddings derived from EEG topographic representations to provide global geometric context for spatial generation, together with a dynamic channel-relation graph that encodes inter-electrode relationships and evolves with temporal dynamics. This design yields a spatially grounded EEG spatial super-resolution framework with consistent performance improvements. Across multiple EEG datasets spanning diverse applications, including SEED/SEED-IV for emotion recognition, PhysioNet motor imagery (MI/MM), and TUSZ for seizure detection, our method achieves substantial gains in generation fidelity and leads to notable improvements in downstream EEG task performance.

%% file: tex/1_intro.tex
\section{Introduction}

\begin{figure}[t]
\centering
\centerline{\includegraphics[width=0.95\columnwidth]{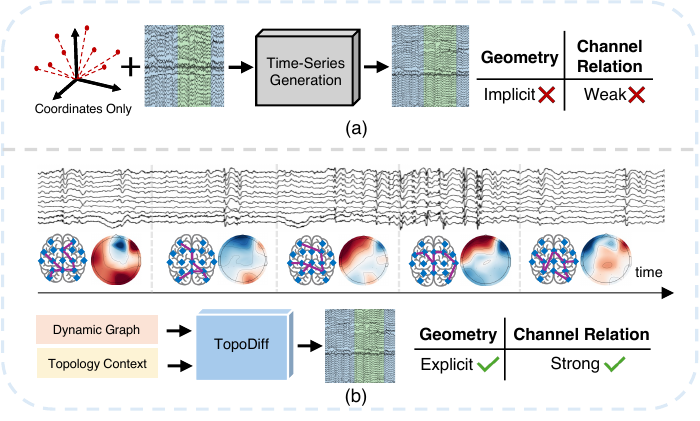}}
\vspace{-10pt}
\caption{Comparison between previous pipelines and the proposed framework. 
(a) Prior methods only implicitly encode spatial geometry through coordinates and positional encoding. 
(b) \textbf{TopoDiff} enables geometry- and relation-aware generation by explicitly incorporating spatial and relational modeling.
}
\vspace{-4mm}
\label{fig:teaser_figure}
\end{figure}

Electroencephalography (EEG) provides a noninvasive window into distributed cortical dynamics of brain and is widely used across affective computing~\cite{zheng2015seed,zheng2018seediv,chen2023affective}, brain–computer interfaces (BCIs)~\cite{goldberger2000physionet, schalk2004mimm,cho2017midata}, and clinical neurophysiology such as epilepsy diagnosis~\cite{herman2015conteeg,waak2023eegicu}.

Despite its broad applicability, standard EEG acquisition typically requires dense electrode placement over the scalp, which introduces hardware constraints and user discomfort, thereby limiting its use in out-of-lab scenarios. In contrast, low-density recording systems, such as wearable EEG devices, enable practical and longitudinal monitoring~\cite{biondi2024athome,cho2025inhomeeeg,vincezo2025wearableeeg,Markov2024eegsleep}. However, these systems rely on sparse electrode layouts that discard fine-grained spatial information, leading to reduced performance in downstream analysis.

This gap has driven increasing interest in EEG spatial super-resolution (EEG SR)~\cite{courellis2016eeg,corley2018deep,svantesson2021virtual}, which aims to generate high-density EEG signals from sparse recordings. More recently, STAD~\cite{wang2024stad} employs a latent diffusion model to generate missing channels, while ESTFormer~\cite{li2025estformer} directly predicts missing signals using a masked autoencoder~\cite{he2022masked}. However, as shown in Figure~\ref{fig:teaser_figure}(a), these methods encode spatial information only implicitly through positional embeddings, which limits their ability to represent rich scalp geometry and to model structured cross-channel dependencies.

These observations motivate explicitly modeling electrode geometry and structured inter-channel relationships, rather than treating spatial positions as passive metadata. In clinical practice, EEG channels are rarely interpreted in isolation; instead, EEG signals are often summarized using \emph{topographic maps} (topoplots), which visualize the spatial distribution of signal amplitudes over the scalp~\cite{Koles1988topo}. Such representations provide a global, geometry-aware view that highlights spatial organization and overall patterns, and naturally generalize across different electrode montages. This perspective suggests a complementary inductive bias for EEG spatial super-resolution: leveraging topographic representations to provide global spatial context. Moreover, electrode distributions can be naturally modeled using graph structures, which enable more structured and relational representations than treating channels as independent signals~\cite{kotoge2025evobrain}.

Building on these insights, as shown in~\cref{fig:teaser_figure}(b), we propose \textbf{TopoDiff}, a spatially grounded EEG SR framework that jointly leverages two structured representations: (i) \textbf{topology-aware image representations} derived from EEG topographic visualizations, which provide global scalp geometry context; and (ii) \textbf{dynamic channel-relation graphs} that model time-varying cross-channel dependencies. Building on recent advances in conditional diffusion models~\cite{esser2024scaling}, we formulate EEG spatial super-resolution as a conditional generation problem, where high-density EEG signals are reconstructed by conditioning on both geometric and relational cues.

Our core contributions are summarized as follows: 
\begin{itemize}
    \item We propose a unified geometry- and relation-aware diffusion framework for EEG spatial super-resolution.
    \item  We introduce \textbf{topology-aware modeling} via EEG topographic image embeddings to capture global scalp geometry alongside \textbf{relation-aware modeling} via channel-relationship graphs to encode structured and time-varying inter-channel dependencies,  enabling generation of fine-grained spatiotemporal EEG patterns. 
    \item Extensive experiments on SEED, SEED-IV, TUSZ, and PhysioNet MI/MM demonstrate improved reconstruction fidelity and stronger downstream performance compared to competitive baselines.
\end{itemize}

%% file: tex/2_related.tex
\section{Related Work}

\textbf{Conditional Diffusion Methods.}
Conditional generation enables data synthesis guided by rich contextual information, ranging from class labels~\cite{peebles2023scalable},  input signals~\cite{wang2024stad} to auxiliary modalities~\cite{esser2024scaling,wan2025wan,wei2025deepseek}. Recently, diffusion models have emerged as a dominant paradigm for conditional generation due to their strong modeling flexibility and high sample quality~\cite{dhariwal2021diffusion,ho2022cfg,saharia2022imagen,rombach2022ldm,esser2024scaling,wan2025wan}. Motivated by this success, we adopt diffusion models for EEG SR and incorporate structured, domain-specific priors as conditioning guidance, rather than relying solely on signal-level inputs.

\textbf{Generative Models for Time Series and EEG Super-Resolution.}
Generative modeling for time series has been widely explored for data augmentation~\cite{shu2025diffeeg}, privacy-preserving synthesis~\cite{tian2024privacy}, and probabilistic forecasting or imputation~\cite{tashiro2021csdi}, with prominent model families including VAEs, GANs~\cite{yoon2019gan}, and, more recently, diffusion models for uncertainty-aware conditional generation~\cite{Rasul2021AutoregressiveDD} (e.g., \textbf{RDPI} improves spatiotemporal imputation via a two-stage deterministic–diffusion framework~\cite{liu2025rdpi}). When applied to EEG, generative models have primarily been used for signal synthesis and data augmentation to mitigate data scarcity, but they typically operate on discrete channel tensors and encode electrode layout only implicitly, without explicitly modeling montage geometry~\cite{shu2025diffeeg,williams2025eeggan}. Recent EEG SR baselines adapt sequence models or diffusion frameworks specifically for spatial upsampling. \textbf{ESTFormer}~\cite{li2025estformer} employs Transformer attention with positional encodings and masking to reconstruct missing channels, while \textbf{STAD}~\cite{wang2024stad} and \textbf{SRGDiff}~\cite{liu2025srgdiff} adopt diffusion-based generation with signal-derived conditioning and step-wise guidance. However, these methods largely rely on positional encodings and learn inter-channel relations implicitly within the model, which can overlook global scalp geometry and structured inter-electrode organization.

\textbf{Downstream EEG Tasks.} EEG has been used for tasks such as emotion recognition~\cite{zheng2015seed}, motor imagery (MI) classification~\cite{schalk2004mimm}, and seizure detection~\cite{herman2015conteeg}, which are commonly adopted to evaluate EEG signal representations. EEG-based emotion recognition~\cite{zheng2015seed, craik2019deep} depends on spatial characteristics such as frontal asymmetry and cross-regional connectivity, potentially impacting multiple regions of the brain, indicating that any degradation in spatial fidelity can affect performance. MI classification~\cite{schirrmeister2017deep, lawhern2018eegnet} decodes limb movement from EEG signals and is highly sensitive to the spatial characterization of sensorimotor rhythms, which low-density montage often fails to capture. Finally, seizure detection provides a clinically grounded downstream task that requires preservation of both localized and distributed pathological patterns. Epileptic activity may manifest as focal or generalized abnormalities~\cite{shah2018temple}, making it extremely important to have an accurate spatial and temporal characterization of EEG signals~\cite{shoeb2009application, roy2019deep}. Collectively, these tasks span affective computing, BCI, and clinical neurophysiology--motivating work in spatially-grounded EEG SR methods.

%% file: tex/3_method.tex
\section{Preliminary}
\subsection{EEG Spatial Super-Resolution}
Let $X^{\mathrm{LR}} \in \mathbb{R}^{C_{\mathrm{LR}} \times T}$ denote a low-resolution (sparse-montage) EEG segment with $C_{\mathrm{LR}}$ channels and $T$ time steps, and let $X^{\mathrm{HR}} \in \mathbb{R}^{C_{\mathrm{HR}} \times T}$ denote the corresponding high-resolution EEG with $C_{\mathrm{HR}}$ channels recorded over the same time window. The EEG spatial super-resolution (EEG SR) task is defined as estimating $X^{\mathrm{SR}} \in \mathbb{R}^{C_{\mathrm{HR}} \times T}$ that approximates $X^{\mathrm{HR}}$, given $X^{\mathrm{LR}}$ as input.
Since the low-resolution channels correspond to a known subset of the high-resolution montage, we define the target as reconstructing the unobserved signals $X^{\mathrm{Unseen}} \in \mathbb{R}^{C_{\mathrm{Unseen}} \times T}$.

\subsection{Denoising Diffusion Models}
Diffusion models learn to generate data by reversing a gradual noising process, which can be interpreted from the perspective of ordinary differential equations (ODEs)~\cite{song2020denoising,chen2018neural,albergo2025stochastic}. During training, a noisy input $z$ is obtained by interpolating between the data distribution and a noise distribution as $z = a x + b \epsilon$, where $\epsilon \sim \mathcal{N}(0, \mathbb{I})$ and $x \sim p(\mathbf{X})$.
For simplicity, following~\cite{li2025back}, we use a linear noise schedule such that, given $t \in (0,1)$, the noisy sample is obtained as $z = t\,x + (1 - t)\epsilon$. We adopt \emph{$x$-prediction}~\cite{li2025back}, where a denoising network $f_\theta$ directly predicts the clean signal at each $t$ as $x_{\mathrm{pred}} = f_\theta\!\left(z,\, t \right)$.
In EEG spatial super-resolution, we formulate the task as a conditional generation problem, where noise is applied only to the unobserved channels, while the observed signals $X^{\mathrm{LR}}$ are provided as conditioning inputs throughout the diffusion process. Specifically, prediction is performed as $X^{\mathrm{Unseen}}_{\mathrm{pred}} = f_\theta\!\left(z,\, X^{\mathrm{LR}},\, t \right)$, and the model is trained with an $\ell_2$ reconstruction loss,
$\mathcal{L}_{\mathrm{diff}} = \mathbb{E}_{x,\epsilon,t}\!\left[\|X^{\mathrm{Unseen}}_{\mathrm{pred}} - x^{\mathrm{Unseen}}\|_2^2\right]$.

At inference time, generation is performed by solving the reverse-time ODE conditioned on $X^{\mathrm{LR}}$. Using $x$-prediction, the corresponding velocity field is given by $v_\theta(z,t) = \frac{x_{\mathrm{pred}} - z}{t}$, yielding the ODE $\frac{\mathrm{d}z}{\mathrm{d}t} = v_\theta(z,t)$. Starting from Gaussian noise, the model iteratively denoises to obtain the generated unseen channels $X^{\mathrm{Unseen}}_{\mathrm{gen}}$.

\begin{figure}[t]
\centering
\centerline{\includegraphics[width=\linewidth]{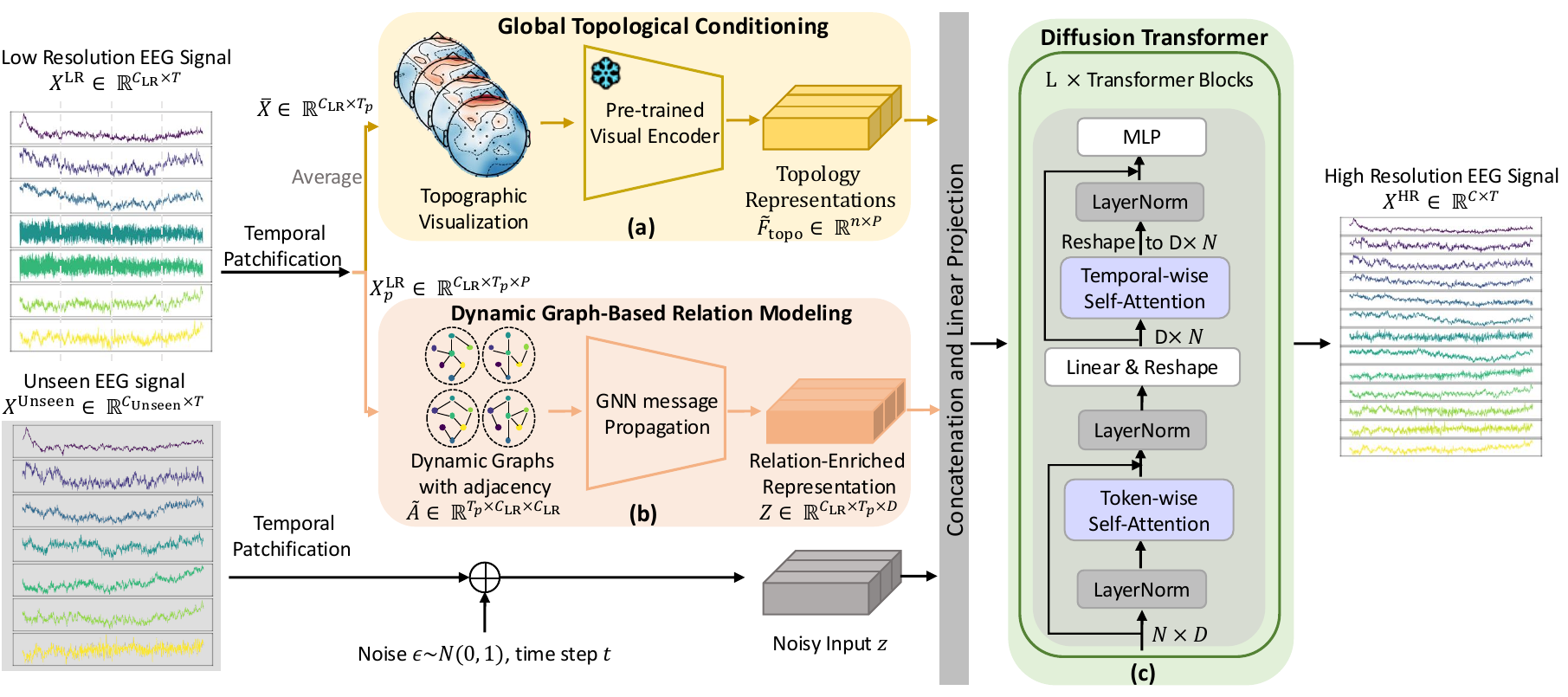}}
\caption{Overview of the \textbf{TopoDiff} framework. (a) Global topological conditioning, where EEG topographic representations are used to provide global scalp geometry and spatial context. (b) Dynamic graph-based inter-channel relation modeling, in which graphs are constructed in a time-varying manner and a GNN is used to enrich channel representations with explicit relational information. (c) The diffusion Transformer architecture employed for EEG spatial super-resolution.
}
\label{fig:spatial_rep}
\end{figure}

\section{TopoDiff Framework}
\label{sec:topodiff}
The core idea of TopoDiff is to integrate geometric and relational inductive biases into the diffusion process in order to ensure spatially and temporally coherent EEG generation. We first describe the diffusion Transformer architecture used for EEG signal generation, and then introduce the global topological conditioning derived from EEG topographic representations, followed by the dynamic graph-based inter-channel relation modeling.

\subsection{Diffusion Transformer Architecture}
As illustrated in Figure~\ref{fig:spatial_rep}, our model adopts a Transformer~\cite{vaswani2017attention} architecture as the denoising generator. Given a low-resolution EEG input $X^{\mathrm{LR}} \in \mathbb{R}^{C_{\mathrm{LR}} \times T}$, we first partition the signal along the temporal dimension into non-overlapping patches, resulting in $X^{\mathrm{LR}}_p \in \mathbb{R}^{C_{\mathrm{LR}} \times T_p \times P}$, where $P$ denotes the temporal patch size and $T_p = T / P$. We then merge the channel and patch-level temporal dimensions to obtain a sequence representation $X^{\mathrm{LR}}_c \in \mathbb{R}^{N \times P}$, where $N = C_{\mathrm{LR}} \times T_p$. This patchification and reshaping strategy enables the Transformer to jointly model spatial (channel-wise) and temporal information within a unified sequence, analogous to spatiotemporal Transformer architectures commonly used in video processing and generation~\cite{wan2025wan}.

We apply the same patchification and reshaping procedure to the noisy input $z$, yielding $z_c \in \mathbb{R}^{N_z \times P}$, where $N_z = C_{\mathrm{Unseen}} \times T_p$. The resulting $z_c$ is concatenated with the conditional representation $X^{\mathrm{LR}}_c$ to form $X_c = [X^{\mathrm{LR}}_c; z_c] \in \mathbb{R}^{(N + N_z) \times P}$. This concatenated sequence is then projected to a latent embedding $X_d \in \mathbb{R}^{(N + N_z) \times D}$ via a linear layer and provided as input to the Transformer, where $D$ denotes the latent feature dimension.

Within each Transformer block, the architecture consists of two attention modules designed to capture spatiotemporal interactions: a token-wise self-attention layer and a temporal-wise self-attention layer. The token-wise self-attention operates over the full set of spatiotemporal tokens of size $N + N_z$, enabling global interactions across channels and time. In contrast, the temporal-wise self-attention first transposes the representation to a temporal-first form $X_t \in \mathbb{R}^{P \times (N + N_z)}$ and performs attention across elements within each temporal patch, facilitating local temporal modeling. After all Transformer blocks, a linear layer projects the features back to the shape of $X_c$, which is then reshaped to the original signal space to produce $X_{\mathrm{pred}} \in \mathbb{R}^{C_{\mathrm{HR}} \times T}$ and supervision is applied only to the unobserved channels.

\subsection{Global Topological Conditioning}
\label{sec:topo_conditioning}
To provide the diffusion model with explicit global scalp geometry information, we incorporate high-level features extracted from EEG topographic representations as conditioning signals. These representations capture coarse-grained spatial organization across the entire scalp, providing global structural context that complements local channel-wise relational modeling. This design can be viewed as a form of cross-modal conditioning, in which image-based representations deterministically derived from EEG signals serve as auxiliary geometric priors to guide time-series generation.

\textbf{Topographic Representation Construction.}
We leverage EEG topographic visualization images~\cite{gramfort2013meg}, which are constructed by interpolating EEG signals onto a two-dimensional projection of the scalp~\cite{perrin1989spherical,Koles1988topo}. Such spatial visualizations are widely used by human experts in EEG analysis to reason about global spatial patterns.

First, low-resolution EEG input $\mathbf{X}^{\mathrm{LR}} \in \mathbb{R}^{C_{\mathrm{LR}} \times T}$ is partitioned along the temporal dimension into non-overlapping patches,
$\mathbf{X}^{\mathrm{LR}}_p \in \mathbb{R}^{C_{\mathrm{LR}} \times T_p \times P}$,
where $P$ denotes the temporal patch size, chosen to be consistent with the Transformer configuration for simplicity.
Within each temporal patch,  a channel-wise temporal average operation is applied,
obtaining $\bar{\mathbf{X}} \in \mathbb{R}^{C_{\mathrm{LR}} \times T_p}$,
which summarizes the dominant signal pattern within each time slot.
Each averaged signal slice is then visualized as a color-coded two-dimensional EEG topographic image
$\mathbf{I}_{\mathrm{topo}} \in \mathbb{R}^{H \times W \times 3}$.

\textbf{Global Feature Extraction and Conditioning.}
A frozen pretrained image feature extractor is applied to the topographic images to obtain a spatial feature map
$\mathbf{F}_{\mathrm{topo}} \in \mathbb{R}^{h \times w \times d}$,
where $h$ and $w$ denote the spatial resolution of the extracted features and $d$ is the feature dimension. We adopt the DINOv3~\cite{simeoni2025dinov3} model due to its strong pixel-level representation capability. The resulting features provide high-level spatial context derived from the EEG topographic representations.

Following conditioning strategies commonly adopted in conditional video generation models~\cite{wan2025wan,yang2024cogvideox}, we flatten the feature map and project it to the same latent dimension as the EEG token representations, yielding
$\tilde{\mathbf{F}}_{\mathrm{topo}} \in \mathbb{R}^{n \times D}$, where $n=h\times w$.
This global topological embedding is concatenated with the EEG token sequence $X_d$ at the beginning of the diffusion Transformer, serving as a global context that guides spatial channel generation throughout the diffusion process.

\subsection{Dynamic Graph-Based Inter-Channel Relation Modeling}
\label{sec:dynamic_graph}
In addition to the diffusion process augmented with global spatial representations, we enhance relational modeling by introducing a dynamic graph neural network (GNN) that operates directly on the observed EEG channels. This module explicitly encodes time-varying cross-channel dependencies, facilitating more coherent multi-channel generation.

\textbf{Group-wise Relation Estimation.}
Given the patchified input $X^{\mathrm{LR}}_p \in \mathbb{R}^{C_{\mathrm{LR}} \times T_p \times P}$, we estimate time-varying inter-channel relations for each temporal group $X^{\mathrm{LR}}_p(:, g, :)$, where $g \in \{0, \ldots, T_p - 1\}$. Specifically, a dense and group-specific relation matrix $A^{(g)} \in \mathbb{R}^{C_{\mathrm{LR}} \times C_{\mathrm{LR}}}$ is computed using cosine similarity:
\begin{equation}
    A^{(g)}_{ij} = \frac{\langle v^{(g)}_i, v^{(g)}_j\rangle}{\|v^{(g)}_i\|\,\|v^{(g)}_j\|},
\end{equation}
where $v^{(g)}_i$ denotes the input of channel $i$ in group $g$. This group-wise relation estimation captures cross-channel dependencies that evolve over time.

\textbf{Graph Construction.}
\label{graph_construction}
Given the group-wise relation matrix $A^{(g)}$, we construct a sparse, relation-aware graph for GNN processing. For each temporal group, the constructed graph nodes correspond to visible EEG electrodes, and spatial locality is imposed using the international 10--20 system~\cite{klem19991020}. Each channel is connected to its $n_s$ nearest neighbors according to electrode coordinates, forming an initial sparse topology. Edge weights are assigned based on the estimated relations by taking the absolute value of $A^{(g)}$ to allow both positive and negative correlations to contribute to relational strength: $\tilde{A}^{(g)}_{ij} = |A^{(g)}_{ij}|.$
To control noise and avoid overly dense graphs, we retain only the top-$k$ weighted edges per node.

\textbf{Relation-Aware Channel Encoding with GNN.}
For each temporal group $g$, we first apply symmetric normalization to the adjacency matrix $\tilde{A}^{(g)}$ following the GCN formulation~\cite{kipf2016semi} to ensure numerical stability:
\begin{equation}
    \bar{A}^{(g)} = \left(D^{(g)}\right)^{-\frac{1}{2}} \hat{A}^{(g)} \left(D^{(g)}\right)^{-\frac{1}{2}},
\end{equation}
where $\hat{A}^{(g)} = \tilde{A}^{(g)} + \mathbb{I},$ and $D^{(g)}_{ii} = \sum_{j=1}^{C_{\mathrm{LR}}} \hat{A}^{(g)}_{ij}.$  

Given the group-wise input $X^{(g)} = X^{\mathrm{LR}}_p(:, g, :) \in \mathbb{R}^{C_{\mathrm{LR}} \times P}$, we first project it to a $D$-dimensional channel embedding $\hat{X}^{(g)} \in \mathbb{R}^{C_{\mathrm{LR}} \times D}$ using a linear layer. We then perform graph-based relation propagation as
\begin{align}
X^{(g)}_{1} &= \sigma\!\left(\bar{A}^{(g)} \hat{X}^{(g)} W_1\right), \\
X^{(g)}_{2} &= \bar{A}^{(g)} X^{(g)}_{1} W_2, \\
Z^{(g)} &= \hat{X}^{(g)} + X^{(g)}_{2},
\end{align}
where $W_1, W_2 \in \mathbb{R}^{D \times D}$ are learnable projection matrices, $\sigma(\cdot)$ denotes a nonlinear activation (e.g., $\tanh$), and the residual connection allows the module to function as a lightweight relational adapter that injects graph-structured interactions into the channel representations. Finally, the resulting relation-enriched features $Z \in \mathbb{R}^{C_{\mathrm{LR}} \times T_p \times D}$ are subsequently fed into the Transformer layers, serving as relation-aware inputs in place of the raw EEG signals.

%% file: tex/4_experiments.tex
\begin{table}[t]
\centering
\setlength{\tabcolsep}{2.8pt}
\renewcommand{\arraystretch}{0.95}
\caption{Comparison of signal fidelity metrics across different datasets. Results are compared against recent baseline methods, including STAD~\cite{wang2024stad}, RDPI~\cite{liu2025rdpi}, and ESTFormer~\cite{li2025estformer}, under $2\times$, $4\times$, and $8\times$ SR settings.
}
{\fontsize{6.5}{7.0}\selectfont 
\resizebox{0.90\textwidth}{!}{
\begin{tabular}{l|ccc|ccc|ccc}
\toprule
& \multicolumn{3}{c|}{2$\times$ SR} & \multicolumn{3}{c|}{4$\times$ SR} & \multicolumn{3}{c}{8$\times$ SR} \\
\cmidrule(lr){2-4}\cmidrule(lr){5-7}\cmidrule(lr){8-10}
Method
& NMSE$\downarrow$ & SNR$\uparrow$ & PCC$\uparrow$
& NMSE$\downarrow$ & SNR$\uparrow$ & PCC$\uparrow$
& NMSE$\downarrow$ & SNR$\uparrow$ & PCC$\uparrow$ \\
\midrule

\multicolumn{10}{l}{\textbf{SEED}} \\
\midrule
STAD~\cite{wang2024stad}
& 0.6780 & 1.689 & 0.530
& 0.7623 & 1.179 & 0.450
& 0.7517 & 0.697 & 0.359 \\
RDPI~\cite{liu2025rdpi}
& 0.5667 & 2.470 & 0.667
& 0.6345 & 1.979 & 0.625
& 0.7238 & 1.406 & 0.544 \\
ESTformer~\cite{li2025estformer}
& 0.5538 & 2.566 & 0.634
& 0.5499 & 2.597 & 0.635
& 0.6585 & 1.815 & 0.551 \\

\midrule
TopoDiff (Ours)
& \textbf{0.4982} & \textbf{3.026} & \textbf{0.674}
& \textbf{0.5447} & \textbf{2.639} & \textbf{0.639}
& \textbf{0.6303} & \textbf{2.004} & \textbf{0.574} \\
\midrule

\multicolumn{10}{l}{\textbf{SEED-IV}} \\
\midrule
STAD~\cite{wang2024stad}
& 0.4763 & 3.221 & 0.720
& 0.6555 & 1.834 & 0.559
& 0.7075 & 1.503 & 0.517 \\
RDPI~\cite{liu2025rdpi}
& 0.3371 & 4.722 & 0.796
& 0.4117 & 3.854 & 0.730
& 0.4915 & 3.085 & 0.687 \\
ESTformer~\cite{li2025estformer}
& 0.2648 & 5.770 & 0.840
& 0.3111 & 5.071 & 0.813
& 0.3912 & 4.076 & 0.765 \\
\midrule
TopoDiff (Ours)
& \textbf{0.2628} & \textbf{5.805} & \textbf{0.843}
& \textbf{0.3091} & \textbf{5.099} & \textbf{0.814}
& \textbf{0.3607} & \textbf{4.429} & \textbf{0.783} \\
\midrule

\multicolumn{10}{l}{\textbf{MI/MM}} \\
\midrule
STAD~\cite{wang2024stad}
& 0.5705 & 2.438 & 0.871
& 0.5985 & 2.229 & 0.875
& 0.6677 & 1.754 & 0.675 \\
RDPI~\cite{liu2025rdpi}
& 0.1686 & 7.651 & 0.872
& 0.2132 & 6.712 & 0.851
& 0.3457 & 4.600 & 0.795 \\
ESTformer~\cite{li2025estformer}
& 0.1788 & 7.477 & 0.902
& 0.2496 & 6.028 & 0.880
& 0.3085 & 5.108 & 0.845 \\
\midrule
TopoDiff (Ours)
& \textbf{0.1591} & \textbf{7.983} & \textbf{0.912}
& \textbf{0.188} & \textbf{7.258} & \textbf{0.896}
& \textbf{0.2326} & \textbf{6.334} & \textbf{0.871} \\
\bottomrule
\end{tabular}
}}
\label{tab:all_datasets_sr}
\vspace{-15pt}
\end{table}

\section{Experiments}

\subsection{Datasets and Preprocessing}
We evaluate on four EEG benchmarks: \textbf{SEED}~\cite{zheng2015seed,duan2013seed2} and \textbf{SEED-IV}~\cite{zheng2018seediv}, \textbf{TUSZ} (TUH EEG Seizure Corpus)~\cite{shah2018temple}, and \textbf{PhysioNet MI/MM}~\cite{schalk2004mimm,goldberger2000physionet}. SEED contains 62-channel EEG from 15 subjects with three emotion labels (positive/neutral/negative) sampled at 200~Hz, while SEED-IV extends this setup to four emotion categories. MI/MM comprises 64-channel EEG sampled at 160~Hz for specific limb movements, and TUSZ provides large-scale clinical scalp EEG for seizure detection with standard 10--20 montages (at least 19 channels) and sampling rates of at least 250~Hz. Collectively, these datasets span affective, motor, and clinical settings, enabling us to verify that the proposed method generalizes across diverse tasks and data collection conditions. For each dataset, we simulate wearable EEG configurations by subsampling channels from the original montage and treat the full montage as ground truth. Electrode coordinates are obtained from the standard international 10--20 system~\cite{klem19991020} and used to construct topology-aware spatial representations. We use subject-based train/test splits to ensure performance reflects cross-subject generalization rather than within-subject memorization.

\textbf{Preprocessing.}
EEG data was first segmented into 4s then resampled to 200Hz (except for MI/MM in which we maintained the original 160Hz) to maintain homogeneity across all datasets. Then, a bandpass filter of 0-75Hz was applied to remove high frequency noise \cite{zheng2015seed}. Finally, the amplitudes of the EEG segments were clamped to \textbf{200uv} in order to prevent extreme outliers and noise from impacting the model negatively.

\begin{wraptable}{r}{0.44\textwidth}
\vspace{-12pt}
\caption{Comparison of signal fidelity metrics on TUSZ at 2$\times$ super-resolution (SR). Metrics include signal fidelity (NMSE, SNR, PCC) and AUROC for seizure detection.}
\label{tab:tusz_2x_sr}
\centering
\small
\setlength{\tabcolsep}{3.5pt}
\renewcommand{\arraystretch}{1.0}
\begin{tabular}{l|ccc|c}
\toprule
Method & NMSE$\downarrow$ & SNR$\uparrow$ & PCC$\uparrow$ & AUROC$\uparrow$\\
\midrule
\rowcolor{gray!20}
Full Data    & -- & -- & -- & 0.878\\
STAD & 0.4519 & 3.450 & 0.634 & 0.540 \\
RDPI & 0.5263 & 2.790 & 0.701 & 0.551\\
ESTformer        & 0.36 & 4.437 & 0.732 & 0.855\\
\midrule
TopoDiff (Ours)             & \textbf{0.3285} & \textbf{4.835} & \textbf{0.760} & \textbf{0.878}\\
\bottomrule
\end{tabular}
\vspace{-5pt}
\end{wraptable}

\textbf{Wearable Subsampling Protocol.} 
We define super-resolution factors of $2\times$, $4\times$, and $8\times$ by selecting sparse subsets of channels from the full montage and reconstructing the full set. Unlike data-driven channel selection techniques that optimize for specific tasks \cite{tam2011performance, gaur2021automatic, pane2018channel}, the SR subsampling was done to ensure that the subsampled channels had adequate scalp coverage. The sampling followed the symmetrical positions commonly used in the 10-5, 10-10, and 10-20 systems~\cite{michel2012towards}, which ensure that the subsampled channels maintained adequate scalp coverage. 
With adequate scalp coverage corresponding to the underlying cortical regions, the selected channels can thus be used to capture the large-scale brain dynamics and distributed cortical activity ~\cite{jurcak200710}.
The channels selected using this paradigm are shown in the Appendix (\cref{tab:channel_downsampling}).

\subsection{Implementation Details}
We implement our method with PyTorch, and all experiments are conducted on four NVIDIA RTX PRO 6000 GPUs. For all models, we use the AdamW optimizer with a learning rate of $5 \times 10^{-4}$ and a weight decay of $0.01$, together with a cosine learning rate scheduler. The diffusion Transformer consists of $L=4$ layers with a hidden dimension of $800$. Input signals are patchified with a patch size of $p=50$ (and $p=40$ for MI/MM to match the slightly different frequency), and all models are trained for 300 epochs. For global topological feature extraction, we adopt DINOv3~\cite{simeoni2025dinov3} and apply it to topographic visualizations constructed from temporally averaged EEG signals. For dynamic graph construction, we utilized $n_s=12, k=4$ for the $2\times$ factor and $n_s=12,k=6$ for $4\times$ and $8\times$ factors (further visualization in the Appendix).

\subsection{Baselines}

\begin{wrapfigure}{r}{0.4\columnwidth}
\vspace{-5mm}
\centering
\includegraphics[width=\linewidth]{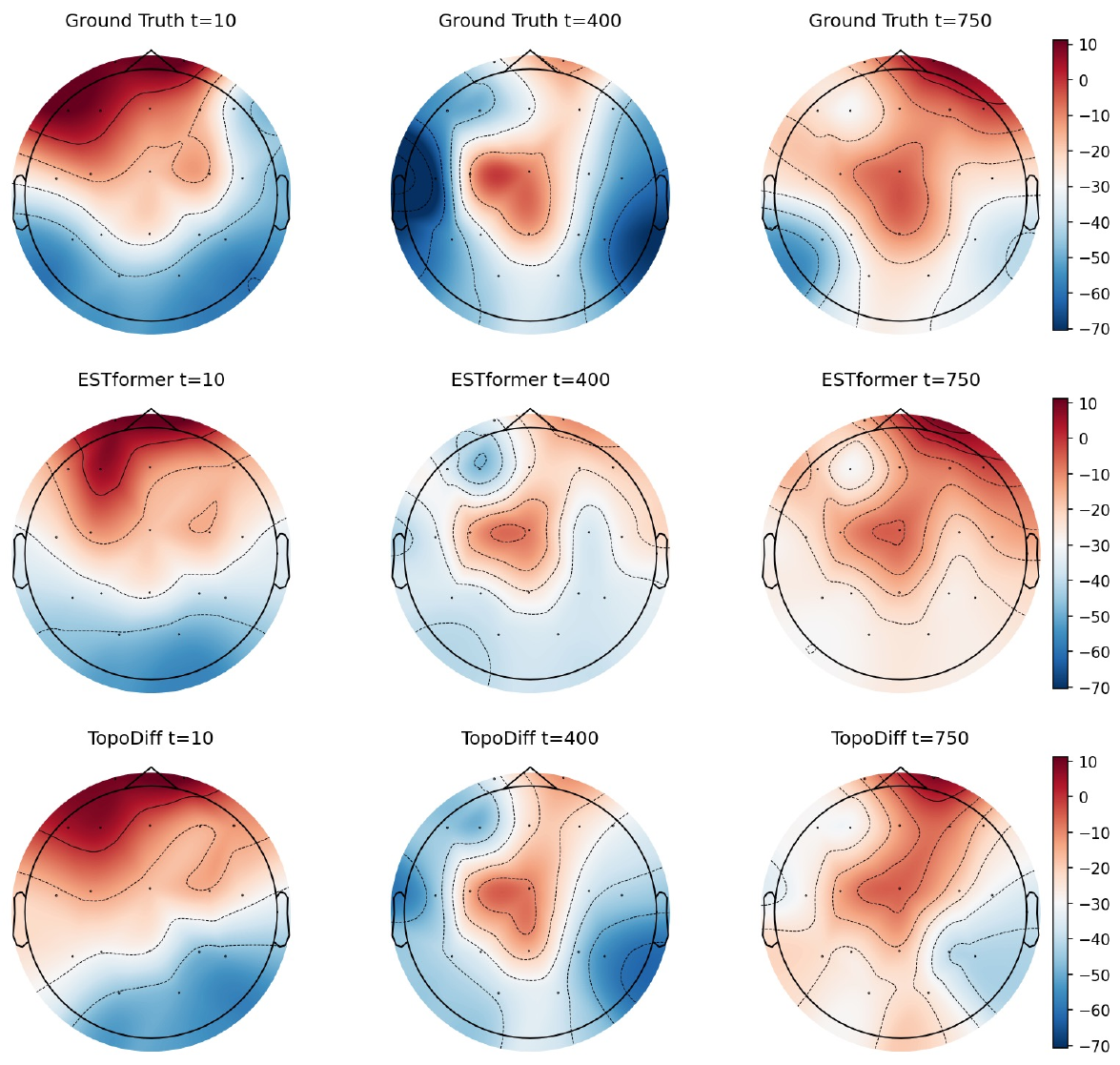}
\caption{Topographical visualization of Ground Truth, ESTformer, and TopoDiff super-resolution outputs for the TUSZ 2$\times$ setting at 3 different time steps.
}
\label{fig:topo_plot}
\vspace{-5mm}
\end{wrapfigure}

We compare against \textbf{ESTformer}~\cite{li2025estformer} and \textbf{STAD}~\cite{wang2024stad} as strong EEG super-resolution baselines, and include \textbf{RDPI}~\cite{liu2025rdpi} as a representative diffusion-based method for general spatiotemporal imputation. All methods are evaluated under identical super-resolution factors ($2\times$, $4\times$, $8\times$) using the same channel selections and identical subject-based train/test splits. We use the authors’ official implementations when available; otherwise, we provide faithful reimplementations and cross-check key architectural and training hyperparameters against the original papers.

For downstream evaluation, we adopt three recent EEG decoding models: \textbf{DMMR}~\cite{wang2024dmmr}, \textbf{3D-CLMI}~\cite{hao20253dclmi}, and \textbf{EvoBrain}~\cite{kotoge2025evobrain}, chosen as competitive, state-of-the-art baselines reported on the corresponding benchmarks. Using fixed downstream architectures isolates the effect of super-resolution quality, allowing us to quantify how improved reconstruction fidelity translates into gains in cross-subject decoding performance. This evaluation reflects common deployment scenarios where sparse wearable EEG is used for downstream inference, and improvements in super-resolution can translate into more reliable and generalizable predictions in real applications. For each of the baselines, we have trained models at each super-resolution factor (utilizing only the newly generated channels) to target the impact of our generated channels and how they perform in these scenarios.

\subsection{Evaluation Metrics}
We report both signal-level reconstruction fidelity and downstream task performance. Signal fidelity is measured using NMSE (normalized mean squared error; lower is better), SNR (signal-to-noise ratio between reconstructed and ground-truth signals; higher is better), and PCC (Pearson correlation coefficient capturing waveform similarity; higher is better). Specifically, we measure the fidelity of the unseen/generated channels for a better comparison on the results of the model. These metrics provide a decoder-agnostic assessment of super-resolution quality that is comparable across datasets and super-resolution factors.

For downstream evaluation, we quantify how reconstruction quality translates to end-task utility: emotion recognition on SEED/SEED-IV, movement (motor imagery) classification on PhysioNet MI/MM, and seizure detection on TUSZ. We report classification accuracy at each super-resolution factor ($2\times$, $4\times$, $8\times$). For TUSZ, since there are only 19 channels, we report metrics $2\times$ only as well as utilize AUROC for downstream task performance to better match metrics typically used for seizure detection~\cite{kotoge2025evobrain}.

\subsection{Main Results}

Table~\ref{tab:all_datasets_sr} and Table~\ref{tab:tusz_2x_sr} summarize signal-level fidelity on SEED/SEED-IV, PhysioNet MI/MM, and TUSZ across multiple super-resolution (SR) factors. Across all datasets and SR levels, our TopoDiff framework achieves the best overall reconstruction quality, consistently improving NMSE, SNR, and PCC relative to strong baselines (ESTformer, RDPI, STAD). The gains are most pronounced in the challenging $8\times$ regime: compared to the best competing method, we improve SNR by \textbf{+10.4\%} on SEED, \textbf{+8.7\%} on SEED-IV, and \textbf{+24.0\%} on MI/MM, alongside substantial NMSE reductions (e.g., MI/MM: 0.2326 vs.\ 0.3085). Lastly, as seen in \cref{tab:tusz_2x_sr}, TopoDiff achieves the lowest NMSE and highest SNR/PCC, with a \textbf{+9.0\%} increase in SNR overall. 

\begin{wraptable}{r}{0.52\textwidth}
\vspace{-12pt}
\caption{Ablation studies on the SEED dataset under the $2\times$ super-resolution setting, including component-wise ablations (Topo: topological representation and Graph: graph-based relation modeling) and analyses of graph construction parameters.
}
\label{tab:seed_ablation}
\centering
\small
\setlength{\tabcolsep}{4.2pt}
\renewcommand{\arraystretch}{1.15}
\begin{tabular}{l|ccc}
\toprule
Method & NMSE$\downarrow$ & SNR$\uparrow$ & PCC$\uparrow$ \\
\midrule
Diffusion Baseline                & 0.5307 & 2.752 & 0.650 \\
Diffusion + Topo                    & 0.5224 & 2.820 & 0.656  \\
Diffusion + Topo + Graph            & 0.4982 & 3.026 & 0.674  \\
\hline
Diffusion + Graph ($n_s$=$12$, top-$k$=$4$)  & 0.5012 & 3.000 & 0.672  \\
Diffusion + Graph ($n_s$=$14$, top-$k$=$8$)   &  0.5185 & 2.852 &  0.659 \\
Diffusion + Graph ($n_s$=$6$, top-$k$=all)   & 0.5062 & 2.957 & 0.669 \\
\bottomrule
\end{tabular}
\vspace{-5pt}
\end{wraptable}

\textbf{Downstream Performance.} Importantly, these signal-level gains translate into improved downstream decoding (Table~\ref{tab:downstream_acc_all_datasets}). On SEED, TopoDiff shows its clearest advantage at 4$\times$ SR, improving accuracy from the best baseline 0.4657 to 0.4902 (\textbf{+5.3\%} relative), and it remains stronger under the most aggressive setting at 8$\times$ SR (0.3890 vs 0.3386, \textbf{+14.9\%}). Similarly, on PhysioNet MI/MM, TopoDiff is \emph{competitive} at 2$\times$ SR (near full-data performance) and yields the best SR accuracy at 8$\times$ SR (0.6155 vs 0.6110, \textbf{+0.7\%}). Overall, the pattern is that improvements become more pronounced as SR becomes harder, suggesting the model is not merely matching waveforms but preserving spatially structured, class-discriminative cues that matter for downstream inference under severe channel sparsity. Additionally, on TUSZ at $2\times$ (Table~\ref{tab:tusz_2x_sr}), we also outperform prior methods on clinical utility, improving AUROC for seizure detection from 0.855 (ESTformer) to 0.878 (which matches the full data performance).

\textbf{Visualization.} Beyond quantitative gains, Figure~\ref{fig:topo_plot} provides qualitative evidence that our reconstructions better preserve the spatial structure of the underlying EEG. The topographic maps show that TopoDiff reproduces regionally consistent patterns that more closely align with the Ground Truth, whereas ESTformer exhibits larger deviations in both the location and magnitude of activity. This closer agreement in spatial distribution suggests that TopoDiff recovers inter-channel relationships more faithfully, yielding signal distribution that are better matched to the original signals. More visualization can be found in the Appendix.

\begin{table}[t]
\centering
\caption{Downstream task accuracy (Acc) across datasets and super-resolution (SR) factors. \colorbox{gray!20}{Full Data} denotes the use of real recordings (super-resolution training targets) for downstream evaluation.}
\label{tab:downstream_acc_all_datasets}
\scriptsize
\setlength{\tabcolsep}{6pt}
\renewcommand{\arraystretch}{1.0}
{
\resizebox{\textwidth}{!}{%
\begin{tabular}{l|ccc|ccc|ccc}
\toprule
\makecell[l]{\textbf{Method}}
& \multicolumn{3}{c|}{\makecell{\textbf{SEED}\\\textbf{(Emotion Recognition)}}}
& \multicolumn{3}{c|}{\makecell{\textbf{SEED-IV}\\\textbf{(Emotion Recognition)}}}
& \multicolumn{3}{c}{\makecell{\textbf{PhysioNet MI/MM}\\\textbf{(Motor Imagery)}}} \\
\cmidrule(lr){1-1}\cmidrule(lr){2-4}\cmidrule(lr){5-7}\cmidrule(lr){8-10}
Settings
& 2$\times$ SR & 4$\times$ SR & 8$\times$ SR
& 2$\times$ SR & 4$\times$ SR & 8$\times$ SR
& 2$\times$ SR & 4$\times$ SR & 8$\times$ SR \\
\midrule
\rowcolor{gray!20}
Full Data        & 0.5685 & 0.5726 & 0.5817 & 0.3068 & 0.3323 & 0.3322 &0.6714 & 0.6924 & 0.6932 \\
STAD             & 0.3523 & 0.4488 & 0.3361 & 0.2116 & 0.2217 & 0.2565 & 0.6428 & 0.6072 & 0.4267 \\
RDPI             & 0.3496 & 0.4657 & 0.3386 & 0.2111 & 0.2151 & 0.2775 & 0.6569 & 0.6478 & 0.6110 \\
ESTformer        & 0.3837 & 0.4309 & 0.3327 & 0.2109 & 0.2256 & 0.2885 & 0.6640 & 0.6424 & 0.5835 \\
\midrule
TopoDiff (Ours)  & \textbf{0.3845} & \textbf{0.4902} & \textbf{0.3890}
                & \textbf{0.2122} & \textbf{0.2272} & \textbf{0.2891}
                & \textbf{0.6701} & \textbf{0.6496} & \textbf{0.6155} \\
\bottomrule
\end{tabular}
}}
\end{table}

\subsection{Ablation Studies}

\textbf{Effect of Graph Relation Modeling.}
Removing graph relation modeling causes a clear drop in reconstruction fidelity. As shown in Table~\ref{tab:seed_ablation}, excluding the graph and relying only on topology conditioning (Diffusion + Topo) reduces SNR by 0.206, and similarly degrades NMSE and PCC. This suggests that the graph module provides essential dynamic inter-channel conditioning, enabling the diffusion model to capture time-varying electrode couplings that are not fully recoverable from global montage context alone.

\begin{wrapfigure}{r}{0.35\columnwidth}
\centering
\centerline{\includegraphics[width=\linewidth]{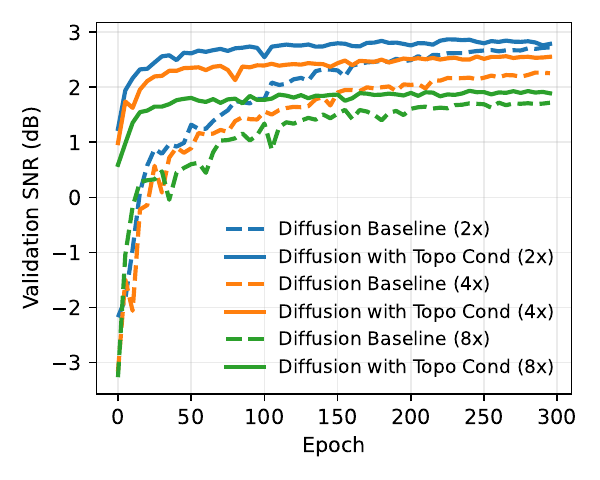}}
\caption{Convergence behavior with topological conditioning on the SEED dataset under $2\times$, $4\times$, and $8\times$ super-resolution settings.
}
\label{fig:topo_training}
\vspace{-5mm}
\end{wrapfigure}

\textbf{Effect of Topological Conditioning.}
As shown in the second row of~\cref{tab:seed_ablation}, under the $2\times$ setting, incorporating topological conditioning improves the SNR by 0.07 and reduces the NMSE by 0.08, demonstrating its effectiveness in facilitating spatial generation. To further analyze how topological conditioning influences learning dynamics, we plot the validation SNR over the course of training in~\cref{fig:topo_training}. It can be observed that, with the global geometric prior provided by the topographic representations, the diffusion model converges significantly faster than the vanilla model. Owing to the global and smoothing characteristics of the topological representation, this conditioning encourages the model to capture spatial geometry early in training, leading to improved convergence behavior. Such behavior is consistent and significant across all super-resolution settings.

\textbf{Impact of Different Sparsity Factors.}
We evaluate several sparsity settings for the two graph-pruning controls in our construction (Table~\ref{tab:seed_ablation}). Overall, graph sparsity materially affects reconstruction quality: overly dense graphs can introduce noisy, weakly relevant couplings, while overly sparse graphs can remove informative spatial interactions, and both regimes degrade signal fidelity. Among the tested configurations, using a smaller neighbor set with a sufficiently rich candidate pool ($n_s{=}12$, top-$k{=}4$; see Sec.~\ref{graph_construction}) yields the best performance compared to denser ($n_s{=}14$, top-$k{=}8$) or less selective ($n_s{=}6$, top-$k{=}$all) alternatives. Intuitively, this setting preserves adaptivity by allowing edge identities and weights to change across patches while still pruning enough, supporting the need to carefully control graph sparsity for stable, spatially consistent super-resolution.

%% file: tex/7_conclusion.tex
\section{Conclusion and Discussion}
Our results highlight that \emph{spatial representation} is a key bottleneck for EEG super-resolution. Rather than treating electrode coordinates as passive positional encodings, our approach conditions a diffusion model on \textbf{global topology-aware topoplot embeddings} and \textbf{channel-relationship graphs}, allowing the denoising process to jointly respect scalp geometry and structured inter-electrode dependencies. Additionally, we highlight the importance of dynamic conditioning for EEG time series which reflects the change in channel relationships across time. In the $2\times$, $4\times$, and $8\times$ super-resolution settings, we observe consistent gains in reconstruction quality and downstream decoding accuracy, indicating that the proposed dual conditioning remains effective as the upsampling task becomes more difficult.

\textbf{Limitations and Future Work.} Several directions could further extend this framework. First, while we use topoplot embeddings as a compact global geometric prior, richer subject-specific spatial cues (e.g., individualized electrode coordinates, head models, or montage calibration) may improve anatomical plausibility and cross-subject transfer, specifically allowing for better and more complete mapping to the brain. Second, our graph conditioning can be expanded to better capture state-dependent connectivity, for example by learning multi-scale or frequency-specific relational graphs and aligning them with known neurophysiological priors. Third, future work can leverage uncertainty for quality control (detecting unreliable reconstructions), active sensing (selecting which additional electrodes would most reduce uncertainty), or downstream-aware training (optimizing SR to maximize task performance). Finally, evaluating robustness under real-world noise, missing channels, and montage shifts enabling unified imputation, denoising, and cross-montage translation would broaden the practical impact of topology- and relation-grounded generative modeling for wearable EEG.

%% file: 99_acknowledgement.tex
\section*{Acknowledgement}
This research was partially funded by the National Institutes of Health (NIH) under award 1R01EB037101-01. The views and conclusions contained in this document are those of the authors and should not be interpreted as representing the official policies, either expressed or implied, of the NIH.

%% file: tex/5_appendix.tex
\section{Implementation Details}
\subsection{Dataset Details}
\textbf{SEED and SEED-IV.}
The SJTU Emotion EEG Dataset (\textbf{SEED})~\cite{zheng2015seed, duan2013seed2} is a standard benchmark for EEG-based emotion recognition. It contains recordings from 15 subjects watching emotion-eliciting video clips labeled with three emotional states: positive, neutral, and negative. EEG is collected using a 62-channel system following the international 10-20 electrode placement standard with a sampling rate of 200 Hz. 
\textbf{SEED-IV}~\cite{zheng2018seediv} extends SEED with four emotion categories (happiness, sadness, fear, and neutrality), while using the same acquisition setup as \textbf{SEED}.

\textbf{Physionet MI/MM.} The PhysioNet MI/MM dataset~\cite{schalk2004mimm, goldberger2000physionet} is a widely used benchmark for MI and motor execution research in EEG-based brain–computer interfaces. It consists EEG recordings from 109 subjects performing or imagining movements of the left hand, right hand, both hands, or both feet in response to visual cues, thus resulting in a three-class classification problem. EEG signals are recorded with a 64-channel international 10–20 montage at a sampling rate of 160 Hz.

\textbf{TUSZ.} TUSZ~\cite{shah2018temple} is a large-scale clinical EEG dataset for seizure detection and classification, consisting of long-term scalp EEG recordings acquired during continuous clinical monitoring. Recordings follow standard clinical 10--20 electrode montages (at least 19 channels) with sampling rates of at least 250~Hz.
\subsection{Channel Selection}

\begin{table*}[t]
\centering
\caption{Channels selected for SR task based on symmetric channel selection.}
\label{tab:channel_downsampling}
\vskip 0.15in
\begin{small} 
\begin{tabularx}{\textwidth}{l p{0.35\textwidth} X X X}
\toprule
\textbf{Dataset} & \textbf{All Channels} & \textbf{2$\times$ SR} & \textbf{4$\times$ SR} & \textbf{8$\times$ SR} \\
\midrule

\makecell[tl]{\textbf{PhysioNet MI/MM} \\ \cite{schalk2004mimm}} & 
\scriptsize Fp1, Fpz, Fp2, AF7, AF3, AFz, AF4, AF8, F7, F5, F3, F1, Fz, F2, F4, F6, F8, FT7, FC5, FC3, FC1, FCz, FC2, FC4, FC6, FT8, T7, C5, C3, C1, Cz, C2, C4, C6, T8, TP7, CP5, CP3, CP1, CPz, CP2, CP4, CP6, TP8, P7, P5, P3, P1, Pz, P2, P4, P6, P8, PO7, PO5, PO3, POz, PO4, PO6, PO8, O1, Oz, O2, Iz & 
\scriptsize Fp1, Fp2, Fz, F3, F4, F7, F8, FC1, FC2, FC5, FC6, Cz, C3, C4, T7, T8, TP7, TP8, CP1, CP2, CP5, CP6, P7, P8, Pz, P3, P4, PO3, PO4, Oz, O1, O2 & 
\scriptsize Fp1, Fp2, F7, F8, T7, T8, F3, F4, C3, C4, P3, P4, P7, P8, O1, O2 & 
\scriptsize F3, F4, C3, C4, P3, P4, O1, O2 \\

\addlinespace[1ex] 

\makecell[tl]{\textbf{SEED/SEEDIV} \\ \cite{zheng2015seed}} & 
\scriptsize Fp1, Fpz, Fp2, AF3, AF4, F7, F5, F3, F1, Fz, F2, F4, F6, F8, FT7, FC5, FC3, FC1, FCz, FC2, FC4, FC6, FT8, T7, C5, C3, C1, Cz, C2, C4, C6, T8, TP7, CP5, CP3, CP1, CPz, CP2, CP4, CP6, TP8, P7, P5, P3, P1, Pz, P2, P4, P6, P8, PO7, PO5, PO3, POz, PO4, PO6, PO8, CB1, O1, Oz, O2, CB2 & 
\scriptsize Fp1, Fp2, AF3, AF4, F7, F3, Fz, F4, F8, FC5, FC1, FC2, FC6, T7, C3, Cz, C4, T8, CP5, CP1, CP2, CP6, P7, P3, Pz, P4, P8, PO3, PO4, O1, Oz, O2 & 
\scriptsize Fp1, Fp2, F7, F8, T7, T8, F3, F4, C3, C4, P3, P4, P7, P8, O1, O2 & 
\scriptsize F3, F4, C3, C4, P3, P4, O1, O2 \\

\addlinespace[1ex]

\makecell[tl]{\textbf{TUSZ} \\ \cite{shah2018temple}} & 
\scriptsize Fp1, Fp2, F3, F4, F7, F8, Fz, C3, C4, Cz, P3, P4, Pz, O1, O2, T3, T4, T5, T6 & 
\scriptsize Fp1, Fp2, F3, F4, C3, C4, O1, O2, Cz & 
\centering -- & 
\centering -- \tabularnewline
\bottomrule
\end{tabularx}
\end{small}
\end{table*}

\begin{figure}[!h]
\centering
\centerline{\includegraphics[width=0.7\columnwidth]{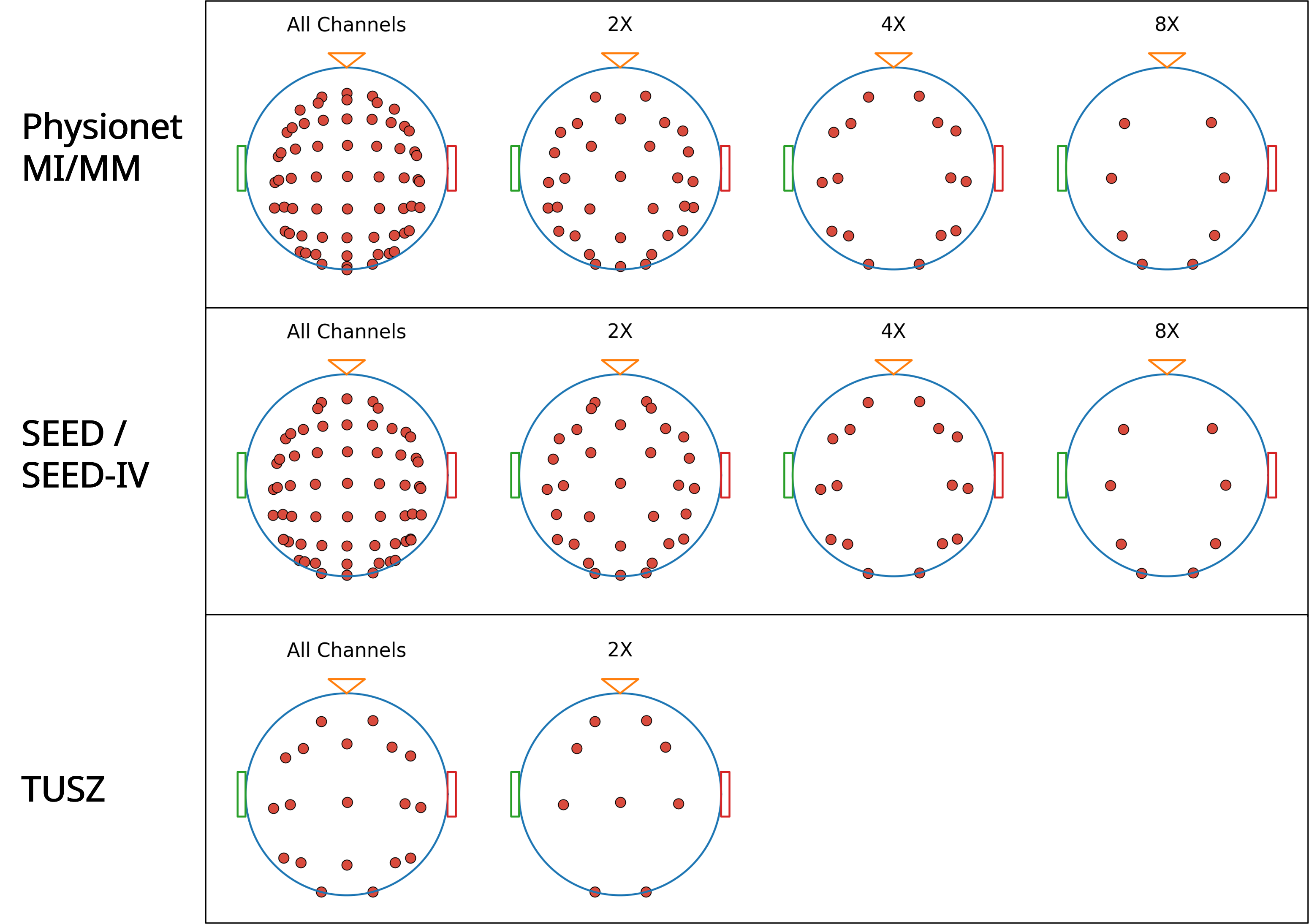}}
\caption{Visualization of EEG channels in full, 2 $\times$ SR, 4 $\times$ SR, and 8 $\times$ SR configurations.}
\label{fig:maps_of_ch}
\end{figure}

Table \ref{tab:channel_downsampling} lists the original channels and the channels selected for SR tasks. 
The channels were selected to ensure that the scalp coverage remains similar to the original electrode configuration. As a result the channels maintain a symmetry.
For TUSZ, only 2 $\times$ SR is considered as it had 19 channels in the full configuration.
Furthermore, Figure \ref{fig:maps_of_ch} visualizes the EEG channels in different configurations, showcasing the symmetric coverage of the scalp.

\subsection{Graph Construction}
\begin{figure}[t]
\centering
\centerline{\includegraphics[width=0.35\columnwidth]{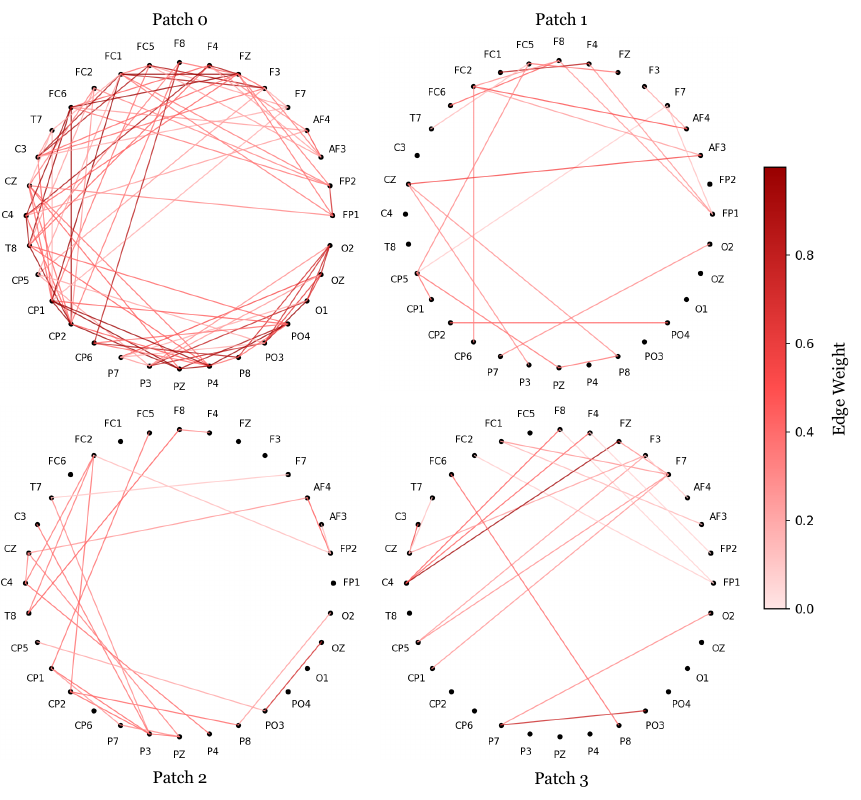}}
\caption{Examples of learned channel-relationship graphs with edge weights visualized by color intensity. The panels show the progression of a sample-specific connectivity pattern over the 10-20 montage at different temporal patches, showing only new edges when compared with the previous patch, highlighting how the relational graph adapts across time while preserving topology-consistent structure.}
\label{fig:graph_weights}
\end{figure}

\begin{figure}[t]
\centering
\centerline{\includegraphics[width=0.35\columnwidth]{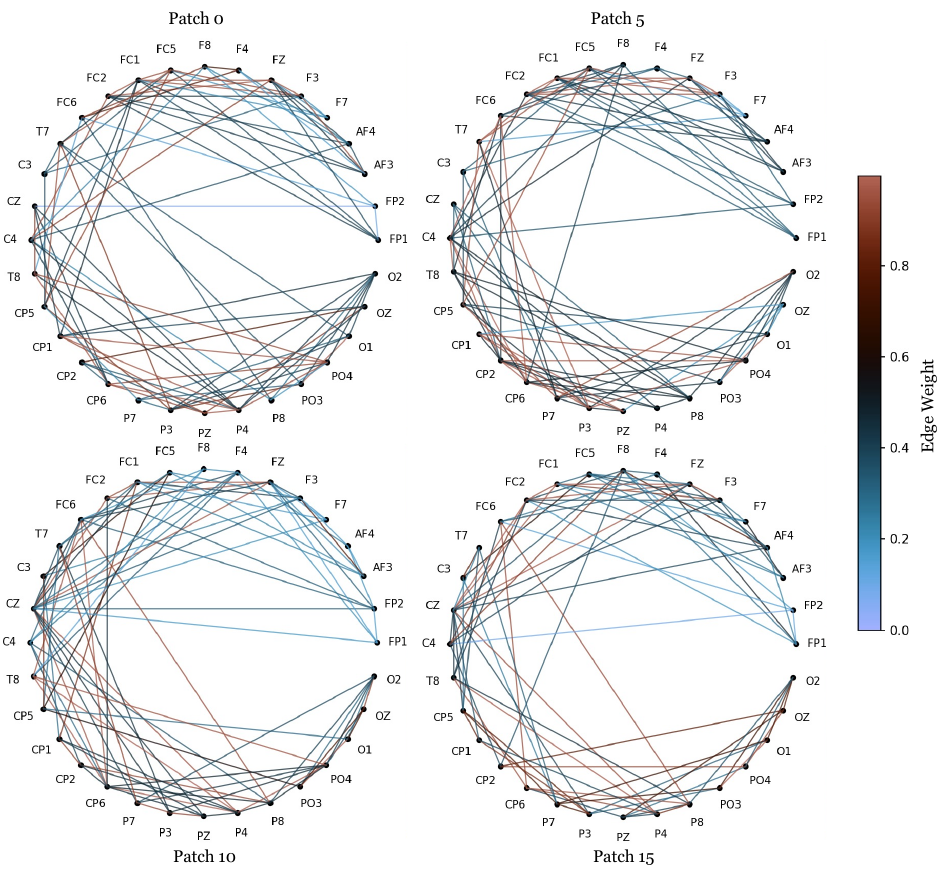}}
\caption{Graph construction over 4 temporal patches with $n_s=12$ and $k=4$ on the SEED 2$\times$ channel setting.
}
\label{fig:graph_construction}
\end{figure}

\cref{fig:graph_weights} and \cref{fig:graph_construction} show visualizations of how the graphs may change over time patches, adding and removing different edges and shifting the weights of existing edges. Both figures map graph constructions for the SEED dataset at the $2\times$ factor, meaning there are 32 visible channels to potentially map connections between. They both demonstrate our strongest parameter setting for SEED $2\times$ which was $n_s=12, k=4$.


\section{Empirical Results}

\begin{figure}[t]
\centering
\centerline{\includegraphics[width=\columnwidth]{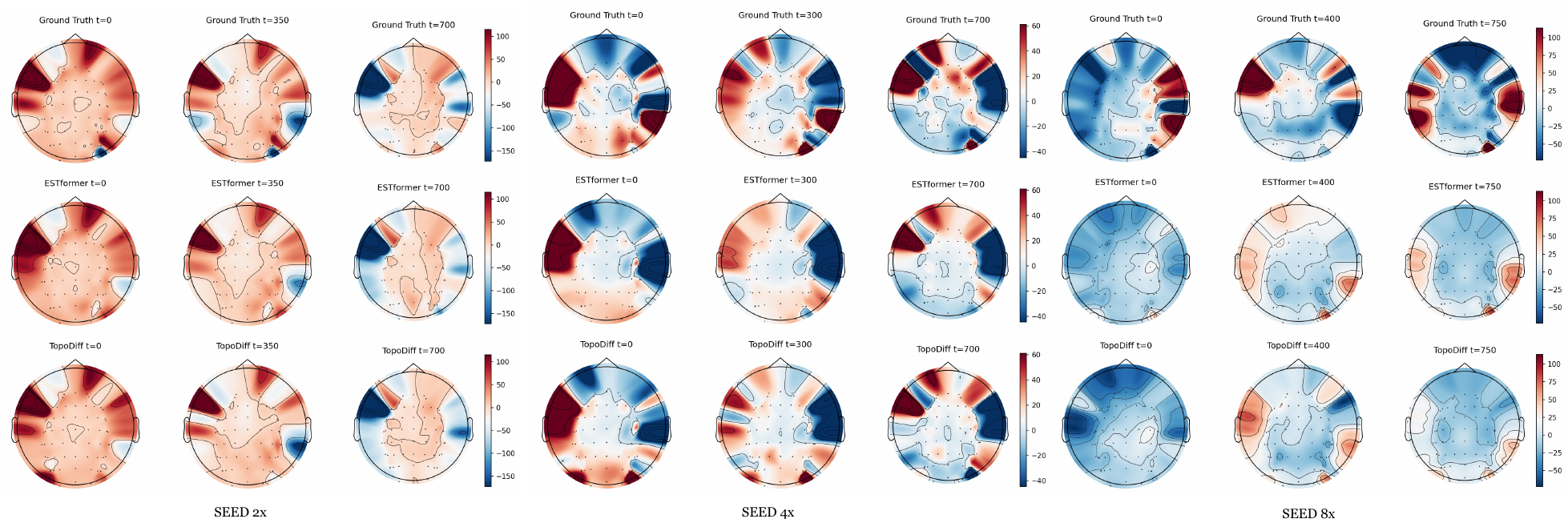}}
\caption{Topographical maps of Ground Truth, ESTformer, and TopoDiff HR outputs for different SEED settings at 3 different time steps.
}
\label{fig:seed_res}
\end{figure}

\begin{figure}[t]
\centering
\centerline{\includegraphics[width=\columnwidth]{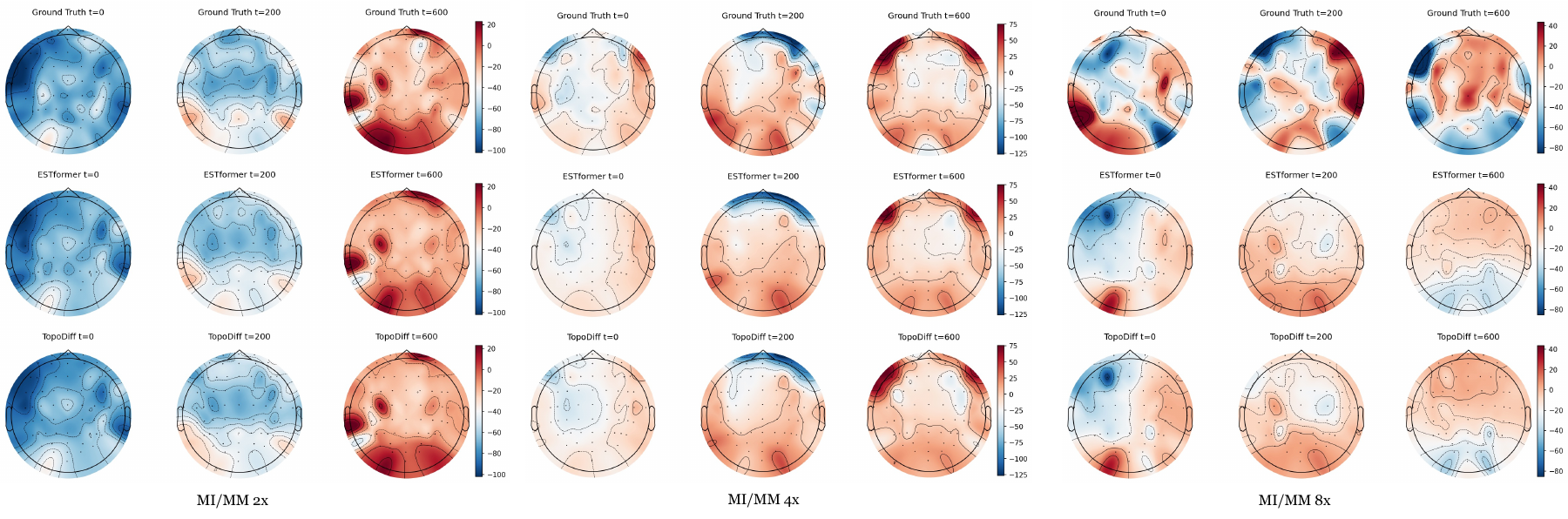}}
\caption{Topographical maps of Ground Truth, ESTformer, and TopoDiff HR outputs for different Physionet MI/MM settings at 3 different time steps.
}
\label{fig:mi_res}
\end{figure}

\cref{fig:seed_res} and \cref{fig:mi_res} further show some results from our super-resolution model when compared at specific time steps to the ground truth and other baseline model generated signals.

